\documentclass{article}
\usepackage{authblk}
\usepackage{spconf,amsmath,graphicx,hyperref}
\usepackage{array}
\usepackage{amsmath,amsfonts}
\usepackage{algorithmic}
\usepackage{graphicx}
\usepackage{textcomp}
\usepackage[dvipsnames]{xcolor}
\usepackage{xspace}
\usepackage{adjustbox}
\usepackage{multicol}
\usepackage{multirow}
\usepackage{booktabs}
\usepackage{pifont}
\usepackage{colortbl}
\makeatletter
\DeclareRobustCommand\onedot{\futurelet\@let@token\@onedot}
\def\@onedot{\ifx\@let@token.\else.\null\fi\xspace}

\def\etal{\emph{et al}\onedot}
\usepackage{hyperref}
\definecolor{citation}{RGB}{79,188,250}
\definecolor{url}{RGB}{255,121,171}
\hypersetup{
    colorlinks=true,
    linkcolor=citation, 
    citecolor=citation, 
    urlcolor=url 
}
\usepackage[capitalize]{cleveref}
\crefname{section}{Sec.}{Secs.}
\Crefname{section}{Section}{Sections}
\Crefname{table}{Table}{Tables}
\crefname{table}{Table}{Tables}
\Crefname{figure}{Figure}{Figures}
\crefname{figure}{Fig.}{Figs.}
\Crefname{equation}{Equation}{Equations}
\crefname{equation}{Eq.}{Eqs.}
\crefname{algocf}{alg.}{algs.}
\Crefname{algocf}{Algorithm}{Algorithms}

\title{ProMist-5K: A Comprehensive Dataset for Digital Emulation of Cinematic Pro-Mist Filter Effects}
\name{
Yingtie Lei$^{1}$ \qquad Zimeng Li$^{2}$ \qquad Chi-Man Pun$^{1}$ \qquad Wangyu Wu$^{3}$ \qquad Junke Yang$^{4}$ \qquad Xuhang Chen$^{1,5^\dagger}$\thanks{{$^\dagger$} Corresponding Author}}
  
\address{ $^{1}$ University of Macau, $^{2}$Shenzhen Polytechnic University, $^{3}$Xi'an Jiaotong-Liverpool University\\$^{4}$University of Illinois Urbana-Champaign, $^{5}$ Huizhou University}

\begin{document}
%
\maketitle
\begin{abstract}
Pro-Mist filters are widely used in cinematography for their ability to create soft halation, lower contrast, and produce a distinctive, atmospheric style. These effects are difficult to reproduce digitally due to the complex behavior of light diffusion. We present ProMist-5K, a dataset designed to support cinematic style emulation. It is built using a physically inspired pipeline in a scene-referred linear space and includes 20,000 high-resolution image pairs across four configurations, covering two filter densities (1/2 and 1/8) and two focal lengths (20mm and 50mm). Unlike general style datasets, ProMist-5K focuses on realistic glow and highlight diffusion effects. Multiple blur layers and carefully tuned weighting are used to model the varying intensity and spread of optical diffusion. The dataset provides a consistent and controllable target domain that supports various image translation models and learning paradigms. Experiments show that the dataset works well across different training settings and helps capture both subtle and strong cinematic appearances. ProMist-5K offers a practical and physically grounded resource for film-inspired image transformation, bridging the gap between digital flexibility and traditional lens aesthetics. The dataset is available at \url{https://www.kaggle.com/datasets/yingtielei/promist5k}.
\end{abstract}
\begin{keywords}
Computational photography, digital cinematography, optical filter emulation
\end{keywords}
\section{Introduction}
The advancement of digital sensors and computational imaging pipelines has reshaped photography and cinematography. Modern cameras provide sharpness, wide dynamic range, and precise tone control, yet digital images often appear overly sharp or ``digital,'' lacking the emotional depth that defines cinematic visuals. By contrast, cinematography is characterized by halation, smooth contrast transitions, and a nostalgic atmosphere~\cite{li2023large}. These qualities are often achieved with optical filters. Neutral density filters regulate exposure~\cite{duma2009neutral}, polarizers enhance saturation and control reflections~\cite{wolff1995polarization}, and color correction filters balance temperature across lighting conditions~\cite{vrhel1994filter}.

Among these tools, Pro-Mist filters are especially valued for their ability to diffuse highlights and reduce contrast while preserving detail. Fine light-scattering particles in the glass scatter bright light into surrounding regions, lowering highlight intensity and lifting shadows. This creates a soft halo, smoother tonal transitions, and a more atmospheric appearance. Such effects enhance depth and emotional tone, aligning with the goals of professional cinematography. However, physical filters require multiple grades and diameters, interrupt on-set efficiency, and fix their effect irreversibly at capture. They also apply uniformly across the frame, offering no selective control.

Digital alternatives promise flexibility but face challenges. Many post-processing tools imitate glow or blur without modeling the underlying light scattering or operating in scene-referred space, often failing to reproduce the balance between halation, contrast attenuation, and detail preservation. Despite the importance of diffusion filters in filmmaking, faithful computational emulation remains underexplored. To address this gap, we introduce ProMist-5K, the first principled framework for digital emulation of Pro-Mist filters. By embedding physically inspired diffusion into a scene-referred pipeline, our method reproduces halation and tonal softness consistent with real filters. This contribution provides a controllable solution for post-production and establishes a foundation for future work in computational cinematography and optical effect emulation.

\section{Related Work}

\subsection{Realistic Cinematic Image Editing}
Research on cinematic rendering has made progress in reproducing film-like qualities such as tonal response, grain, and depth of field, but the simulation of halation remains largely overlooked. Gong~\etal~\cite{gong2024film} proposed Film-GAN to reproduce tonal response and grain, and Li~\etal~\cite{li2023large} built FilmSet, a dataset of 5,000 images with calibrated film styles for tone and color emulation. Other studies have extended this direction to perceptual attributes. Ameur~\etal~\cite{ameur2023style} introduced FilmGrainStyle740k for grain synthesis, Ignatov~\etal~\cite{ignatov2020rendering} released the EBB! dataset for bokeh emulation, and Xian~\etal~\cite{conde2023lens} presented BETD with synthetic blur variations. These works have provided valuable resources for modeling film-like appearance, yet they focus mainly on tone, grain, and blur. None of them capture halation, which filmmakers regard as essential for conveying cinematic softness and atmosphere. Addressing this missing dimension is central to our work, as we aim to provide a principled framework for the emulation of Pro-Mist filters that reproduces this key optical effect.
\begin{figure}[t]
    \centering
    \includegraphics[width=\linewidth]{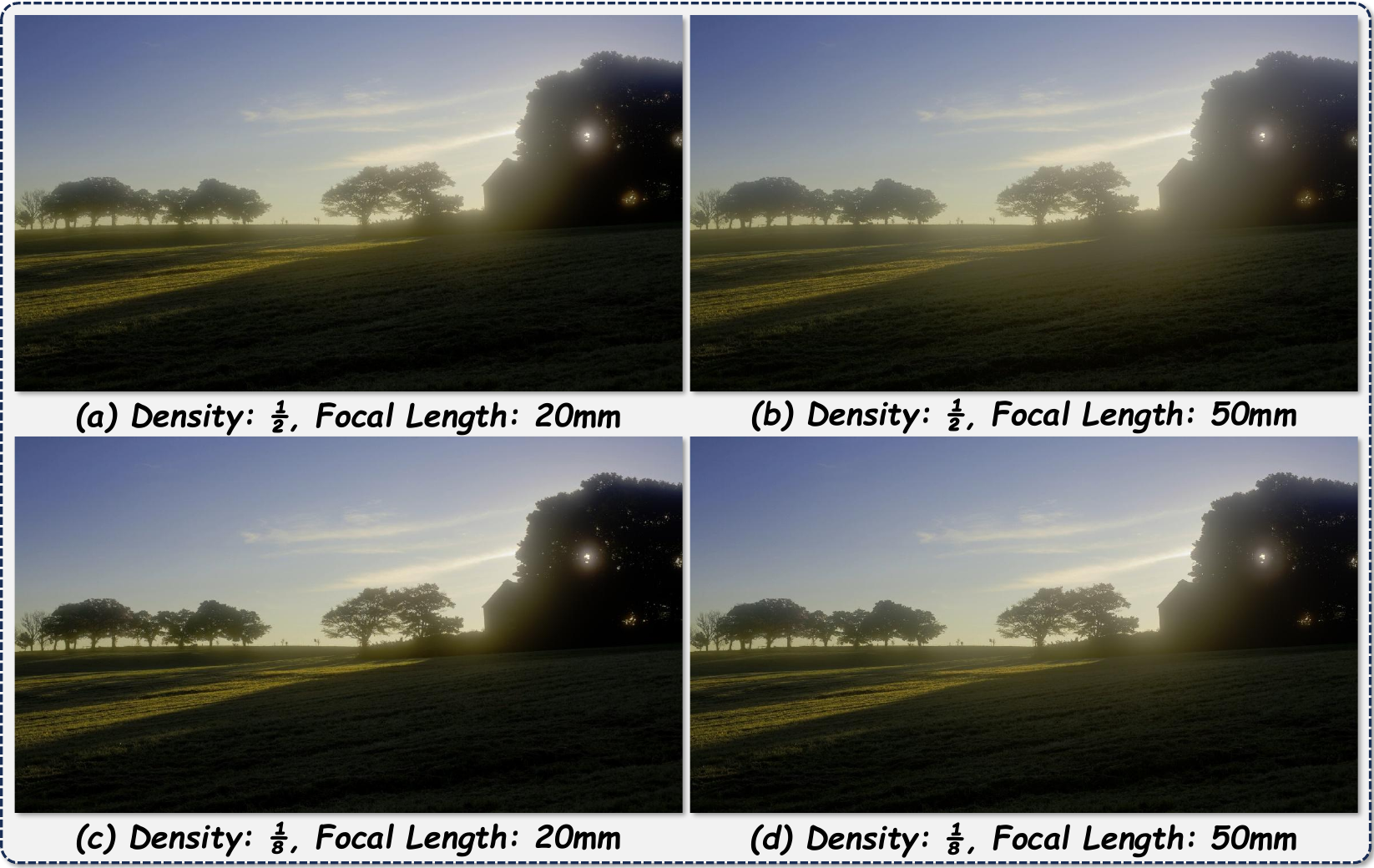}
    \caption{Comparison of effects under different densities and focal lengths. At a fixed density, increasing the focal length from (a) to (b) intensifies highlight diffusion. At a fixed focal length, reducing the density from (a) to (c) weakens the softening effect. (d) shows that low density combined with longer focal length maintains a subtle yet visible cinematic atmosphere.
    } 
    \label{fig:example}
\end{figure}
\subsection{Artistic Image Stylization}
Artistic image stylization aims to transfer aesthetic qualities from artworks to photographs, focusing on creative transformation rather than real lens behavior. Early work by Gatys~\etal~\cite{gatys2016image} demonstrated that neural networks can separate content and style through feature statistics, which enabled style transfer from paintings to images. This line of research was extended by AdaIN~\cite{huang2017arbitrary} for real-time arbitrary style transfer and CycleGAN~\cite{zhu2017unpaired} for unpaired translation. More recently, diffusion-based methods such as InST~\cite{zhang2023inversion} and StyleDiffusion~\cite{wang2023stylediffusion} have improved consistency and preserved stylistic details.
\begin{table}[b]
\centering
\caption{Summary of the ProMist-5K Dataset}
\label{tab:promist-config}
\begin{adjustbox}{width=\linewidth}
\begin{tabular}{@{}cccc@{}}
\toprule
\textbf{Configuration} & \textbf{Training} & \textbf{Testing} & \textbf{ Visual Effect Description} \\ \midrule
1/8 @ 20mm & 4,500 & 500 & Subtle diffusion with high detail retention \\
1/8 @ 50mm & 4,500 & 500 & Mild halation with balanced contrast \\
1/2 @ 20mm & 4,500 & 500 & Strong bloom with softened contrast \\
1/2 @ 50mm & 4,500 & 500 & Heavy diffusion with dramatic cinematic glow \\ \midrule
Total & 18,000 & 2,000 & Comprehensive range of cinematic effects \\ \bottomrule
\end{tabular}
\end{adjustbox}
\end{table}Several datasets support this task. WikiArt~\cite{phillips2011wiki} is a widely used resource of artworks from diverse movements, and MS-COCO~\cite{lin2014microsoft} provides content supervision. BAM!\cite{wilber2017bam} adds labeled artworks from modern design, while StyleGallery\cite{gao2024styleshot} integrates WikiArt, JourneyDB~\cite{sun2023journeydb}, and LAION-Aesthetics~\cite{schuhmann2022laion} to cover a broad range of styles. Other benchmarks, such as BAID~\cite{yi2023towards} and APDDv2~\cite{jin2024apddv}, include structured annotations to support evaluation. Although these datasets advance stylization research, they do not capture the physics-based optical properties of cinematography, which highlights the need for resources that reflect real lens effects.

\section{ProMist-5K Dataset}
\subsection{Dataset Description}
We present ProMist-5K, a paired dataset designed for research in physically grounded cinematic image translation. It extends the Adobe-MIT FiveK dataset~\cite{bychkovsky2011learning} and reproduces the distinctive visual behavior of Pro-Mist filters under realistic conditions. The dataset covers four filter settings that combine two diffusion strengths (1/2 and 1/8) with two focal lengths (20mm and 50mm). As summarized in~\cref{tab:promist-config}, each configuration contains 4,500 training pairs and 500 testing pairs, resulting in a total of 20,000 high-resolution samples. Each pair consists of an emulated Pro-Mist image and its corresponding unfiltered original.

The interaction between density and focal length produces systematic differences in visual appearance, as shown in~\cref{fig:example}. At a fixed density of 1/2, increasing the focal length from 20mm to 50mm strengthens highlight diffusion and halation. At a constant focal length of 20mm, reducing the density from 1/2 to 1/8 weakens the softening while preserving more structural detail. Even with a low density of 1/8, a longer focal length of 50mm still introduces a subtle but noticeable cinematic glow.

\begin{figure}[h]
    \centering
    \includegraphics[width=\linewidth]{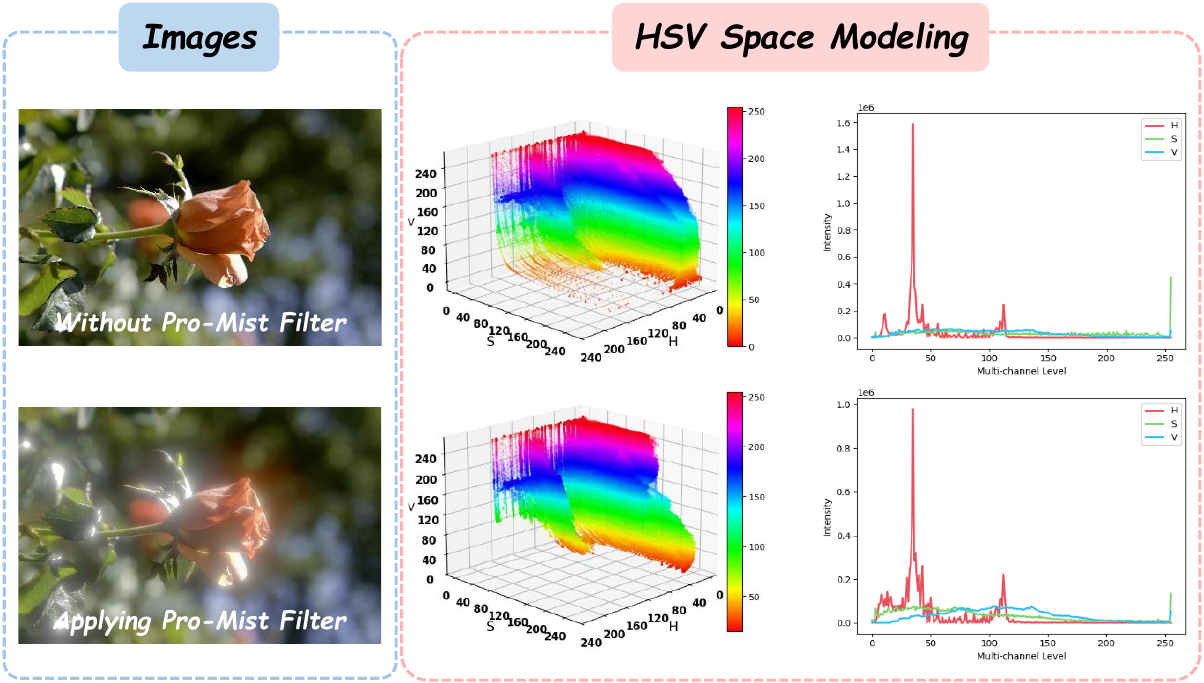}
    \caption{Pro-Mist filter effect (1/2@20mm): Original vs. filtered image with their 3D HSV space models and 1D histograms. Altered HSV distributions illustrate highlight suppression, contrast reduction, and desaturation.
    } 
    \label{fig:example2}
\end{figure}

To further illustrate the filter’s influence,~\cref{fig:example2} analyzes the 1/2@20mm setting in HSV color space. The diffusion of light reduces brightness peaks, compresses dynamic range, and smooths tonal transitions. It also lowers saturation, shifting distributions toward weaker color intensity. Hue remains relatively stable, but the reduction of peak values suggests a softening of dominant tones. These quantitative changes confirm how Pro-Mist filters balance highlight suppression, contrast reduction, and color desaturation, reinforcing their distinctive filmic aesthetic. This analysis highlights the value of ProMist-5K as a reliable benchmark for studying cinematic diffusion effects.

\subsection{Dataset Construction}
\begin{figure}[b]
    \centering
    \includegraphics[width=\linewidth]{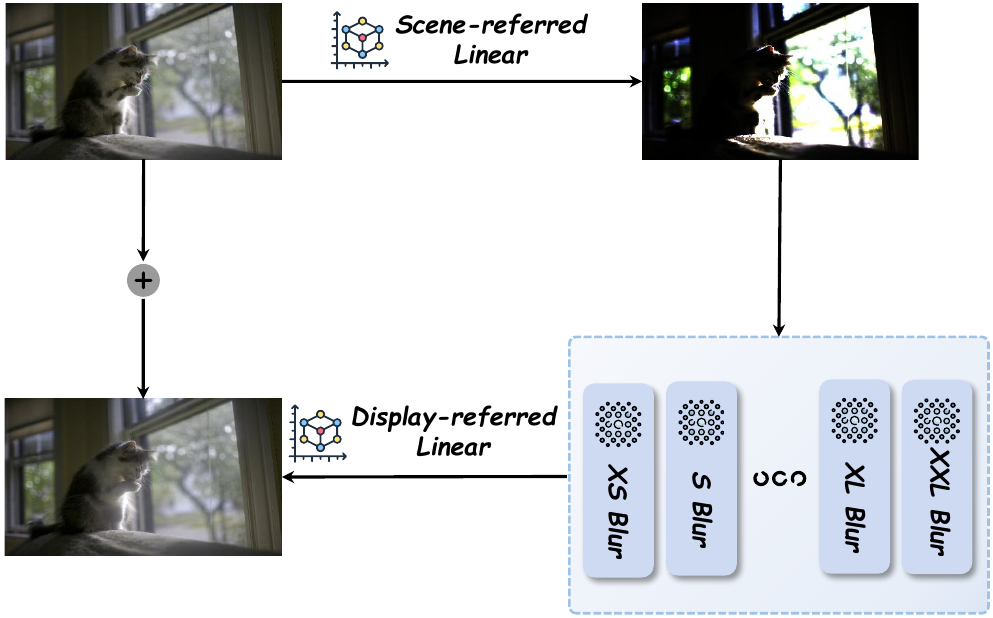}
    \caption{Pro-Mist Filter Emulation Pipeline.
    } 
    \label{fig:construction}
\end{figure}
\begin{figure}[h]
    \centering
    \includegraphics[width=\linewidth]{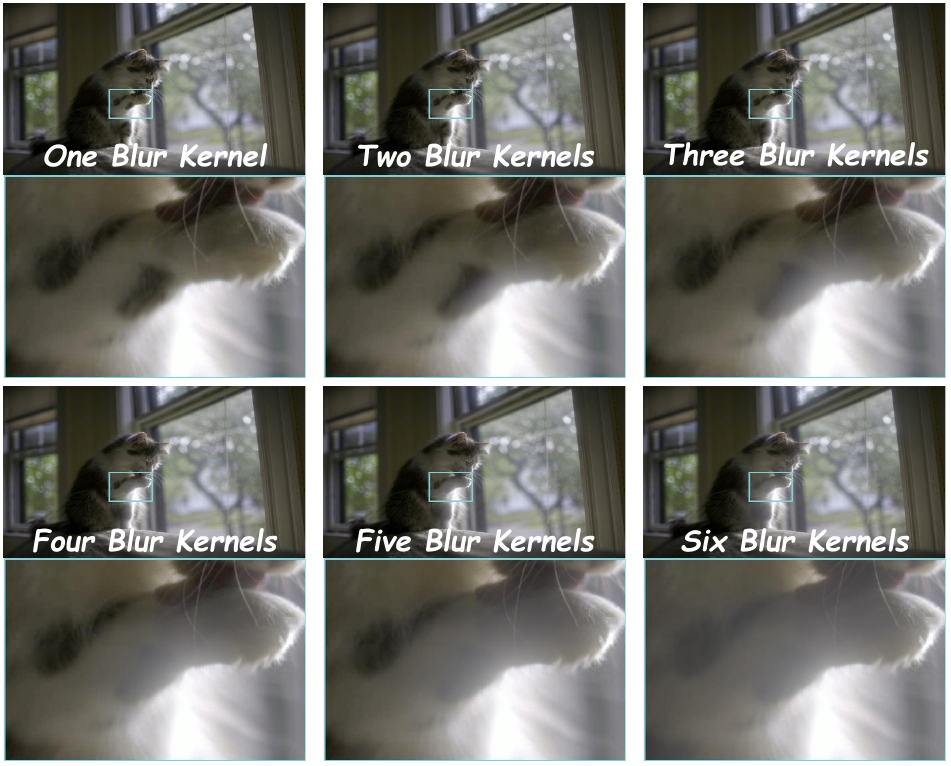}
    \caption{Contribution of Multi-Scale Blur Layers to Pro-Mist Emulation. Increasing the number of blur layers improves the fidelity of halation rendering.
    } 
    \label{fig:blur}
\end{figure}
We construct the ProMist-5K dataset through an emulation pipeline designed to reproduce the optical behavior of physical Pro-Mist filters. As shown in~\cref{fig:construction}, the pipeline operates in a scene-referred linear color space, where pixel values are proportional to real-world light intensities. This setting enables accurate modeling of how bright regions scatter light into surrounding areas, producing the halation and glow typical of diffusion filters. Each image is first converted to the linear space by removing gamma encoding, which allows subsequent operations to simulate physical light behavior more faithfully.

To approximate diffusion, we apply six Gaussian blur layers with progressively larger kernels. These layers represent different scattering scales produced by particles inside a real filter. The perceived filter density is controlled by adjusting the blending weights of these blurred layers. In the 1/2 density setting, larger-radius layers dominate and produce stronger halos. In the 1/8 setting, smaller-radius layers are emphasized, leading to softer diffusion and better preservation of structure.~\cref{fig:blur} demonstrates how increasing the number of blur layers improves spatial richness and tonal smoothness, yielding a more natural result. The pipeline also accounts for focal length. At the same density, images captured at longer focal lengths such as 50mm are simulated with larger blur radii than those at 20mm. This adjustment matches the more pronounced and spatially concentrated halation of telephoto lenses. After applying the appropriate blur weighting and kernel scaling, the blurred layers are combined with the original scene-referred image. The result is then tone-mapped to a display-referred format. Each stylized image is paired with its unfiltered counterpart, creating a supervised training pair. In total, the dataset contains 20,000 pairs evenly distributed across four configurations. These samples form a benchmark for tasks such as highlight diffusion, halation emulation, and broader cinematic enhancement.

\section{Baseline Evaluation}
\subsection{Baselines and Experimental Setup}
\begin{figure*}[h]
    \centering
    \includegraphics[width=\linewidth]{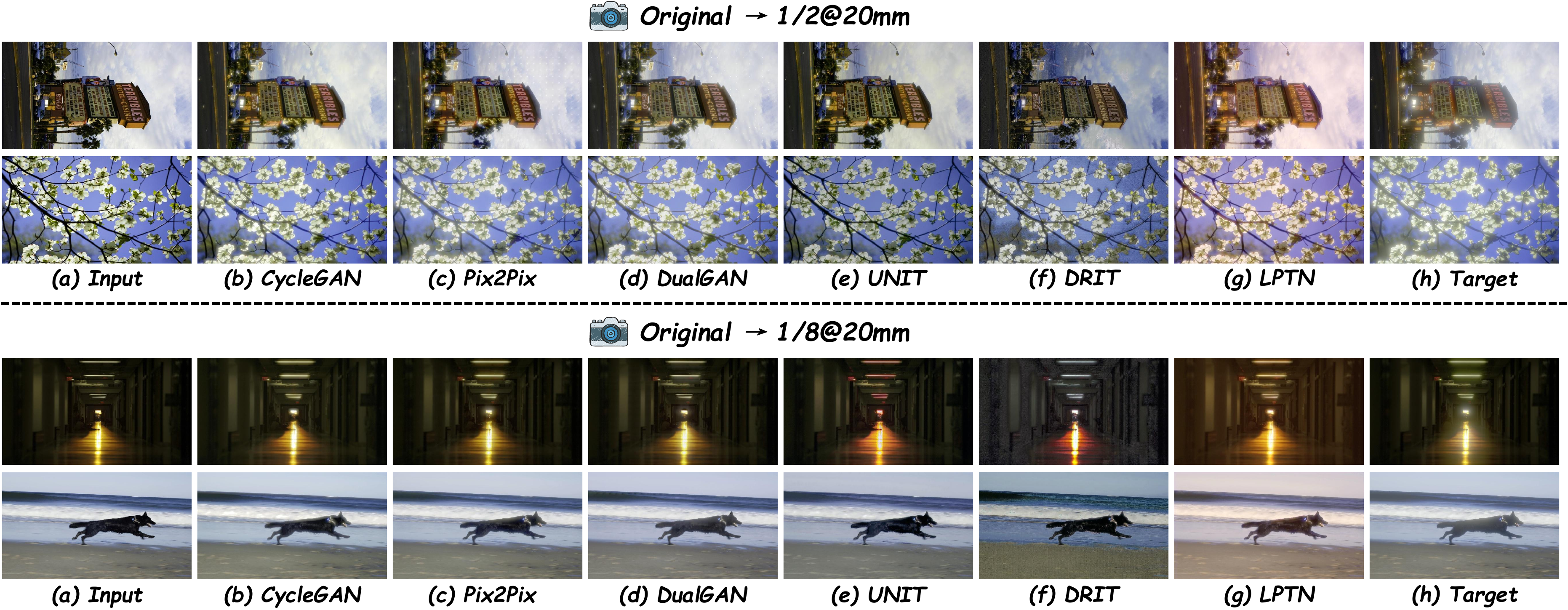}
    \caption{Qualitative comparison of different translation models on two representative Pro-Mist emulation tasks (Original $\rightarrow$ 1/2@20mm and Original $\rightarrow$ 1/8@20mm).}

    \label{fig:visual}
\end{figure*}
Image-to-image translation seeks to convert an image from one domain to another while preserving structure and semantics. In our case, the goal is to transform a clean digital image into a stylized version that reproduces the visual effects of a Pro-Mist filter. ProMist-5K is particularly suitable for this task because it provides high-resolution paired images that include both the unfiltered input and its emulated counterpart with halation and highlight diffusion. This design supports paired and unpaired translation approaches and creates a consistent domain grounded in physical principles, which helps models learn in a stable and generalizable way.

To assess the dataset, we evaluate several representative translation models: CycleGAN~\cite{zhu2017unpaired}, Pix2Pix~\cite{isola2017image}, DualGAN~\cite{yi2017dualgan}, UNIT~\cite{liu2017unsupervised}, DRIT~\cite{lee2018diverse}, and LPTN~\cite{liang2021high}. These baselines cover both paired and unpaired training and are widely used in image translation research. All experiments are run on an NVIDIA RTX A6000 GPU using the official implementations with default settings. Each model is trained for 200 epochs with a batch size of 8. We report results on two representative tasks: translating original images to the 1/2@20mm version and to the 1/8@20mm version. These two settings differ in the strength and spread of highlight diffusion, which allows us to test performance on both pronounced and subtle emulation cases.
\begin{table}[t]
\centering
\caption{Quantitative evaluation results on two Pro-Mist emulation tasks (Original $\rightarrow$ 1/2@20mm and Original $\rightarrow$ 1/8@20mm) using PSNR, SSIM, LPIPS, and FID. The best score for each metric is shown in \textcolor{red}{red}, and the second-best is shown in \textcolor{blue}{blue}.}

\label{tab:evaluation}
\begin{adjustbox}{width=\linewidth}
\begin{tabular}{l|cccc|cccc}
\toprule
\multirow{2}{*}{Methods} & \multicolumn{4}{c|}{Original~$\rightarrow$ 1/2@20mm} & \multicolumn{4}{c}{Original~$\rightarrow$ 1/8@20mm} \\
\cmidrule(lr){2-5} \cmidrule(lr){6-9}
 & PSNR↑ & SSIM↑ & LPIPS↓ & FID↓ & PSNR↑ & SSIM↑ & LPIPS↓ & FID↓ \\
\midrule
CycleGAN~\cite{zhu2017unpaired} & 24.6045 & 0.9114 & 0.0717 & 57.9102 & 28.7285 & 0.9478 & 0.0329 & 31.2321 \\
Pix2Pix~\cite{isola2017image} & \textcolor{red}{26.3893} & \textcolor{blue}{0.9154} & \textcolor{red}{0.0618} & \textcolor{red}{49.1602} & \textcolor{red}{30.1434} & \textcolor{red}{0.9592} & \textcolor{red}{0.0257} & \textcolor{red}{25.1520} \\
DualGAN~\cite{yi2017dualgan} & \textcolor{blue}{25.9989} & \textcolor{red}{0.9252} & \textcolor{blue}{0.0664} & \textcolor{blue}{54.6365} & \textcolor{blue}{29.4912} & \textcolor{blue}{0.9584} & \textcolor{blue}{0.0310} & \textcolor{blue}{30.5926} \\
UNIT~\cite{liu2017unsupervised} & 20.6195 & 0.8648 & 0.1070 & 71.4165 & 24.0747 & 0.9256 & 0.0511 & 43.2495 \\
DRIT~\cite{lee2018diverse} & 18.4447 & 0.7752 & 0.1674 & 86.1399 & 19.4654 & 0.8094 & 0.1244 & 68.3024 \\
LPTN~\cite{liang2021high} & 21.3794 & 0.8870 & 0.1003 & 64.8371 & 21.6460 & 0.9112 & 0.0559 & 42.0154 \\
\bottomrule
\end{tabular}
\end{adjustbox}
\end{table}

\subsection{Results and Analysis}
\cref{tab:evaluation} shows the overall performance of different models under two filter settings. Paired model like Pix2Pix perform more reliably, especially in capturing soft halation and preserving image structure. Unpaired models like CycleGAN and DualGAN also show reasonable results, which suggests that the dataset can be used in both supervised and unsupervised settings without major limitations.~\cref{fig:visual} provides visual examples that reflect these differences. For the 1/2@20mm task, models need to reproduce stronger glow and wider light spread, which is more difficult to model. Paired models produce smoother and more natural-looking transitions, while unpaired models may introduce blur artifacts or color inconsistencies around highlights. In the 1/8@20mm case, where the diffusion is more subtle, most models generate cleaner results with better detail preservation.

These results confirm that ProMist-5K supports a wide range of image translation tasks. It provides a consistent setting for evaluating both subtle and strong diffusion effects, and allows fair comparison across different model types.
\section{Conclusion}
This work introduces ProMist-5K, a dataset for emulating Pro-Mist filter effects that are widely used in cinematography to create softness and atmosphere. Unlike methods that apply generic artistic filters, our dataset follows physically consistent principles and reproduces optical characteristics such as halation, glow diffusion, and contrast reduction with control and accuracy. The scene-referred emulation pipeline models lens behavior under different densities and focal lengths, enabling fine-grained reproduction of subtle cues. Comprehensive evaluation demonstrates that ProMist-5K offers a strong foundation for analyzing and learning cinematic diffusion effects. It supports flexible experimentation with data-driven approaches and addresses a gap in existing resources by focusing on lens behaviors that are central to storytelling and emotional tone. We expect this dataset to facilitate future research in stylization, post-production, and computational cinematography, contributing to more expressive and controllable image creation.

\section{Acknowledgment}
This work was supported in part by the Science and Technology Development Fund, Macau SAR, under Grant 0193/2023/RIA3 and 0079/2025/AFJ, in part by the University of Macau under Grant MYRG-GRG2024-00065-FST-UMDF, and in part by the Guangdong Basic and Applied Basic Research Foundation (Grant No. 2024A1515140010).

\bibliographystyle{IEEEbib}
\bibliography{refs}

\end{document}